\documentclass[runningheads]{llncs}
\usepackage[T1]{fontenc}
\usepackage{graphicx}
\usepackage{amsmath,amssymb} 
\usepackage{color}

\usepackage{algorithm}
\usepackage{algorithmic}

\usepackage{amsmath} 
\usepackage{xcolor} 

\usepackage{newfloat}
\usepackage{listings}
\usepackage{multirow}
\usepackage{xspace}
\usepackage{caption}
\def\ie{\emph{i.e.}}
\def\eg{\emph{e.g.}}

\newcommand{\etal}{\textit{et al}.}
\newcommand{\mini}{\textit{mini}}

\begin{document}

\newcommand{\ours}{CRML}

\title{Few-shot Metric Learning: \\ Online Adaptation of Embedding for Retrieval}
\titlerunning{Few-shot Metric Learning}
%
\author{Deunsol Jung \and
Dahyun Kang \and
Suha Kwak \and 
Minsu Cho}
\authorrunning{D. Jung et al.}
%
\institute{Pohang University of Science and Technology (POSTECH), South Korea \\
\email{\{deunsol.jung,dahyun.kang,suha.kwak,mscho\}@postech.ac.kr}}
\maketitle              
%

\begin{abstract}
Metric learning aims to build a distance metric typically by learning an effective embedding function that maps similar objects into nearby points in its embedding space. Despite recent advances in deep metric learning, it remains challenging for the learned metric to generalize to unseen classes with a substantial domain gap. 
To tackle the issue, we explore a new problem of few-shot metric learning that aims to adapt the embedding function to the target domain with only a few annotated data. We introduce three few-shot metric learning baselines and propose the \emph{Channel-Rectifier Meta-Learning} (CRML), which effectively adapts the metric space online by adjusting channels of intermediate layers. 
Experimental analyses on \textit{mini}ImageNet, CUB-200-2011, MPII, as well as a new dataset, \textit{mini}DeepFashion, demonstrate that our method consistently improves the learned metric by adapting it to target classes and achieves a greater gain in image retrieval when the domain gap from the source classes is larger.
    
\end{abstract}


\section{Introduction}

The ability of measuring a reliable distance between objects is crucial for a variety of problems in the fields of artificial intelligence. 
Metric learning aims to learn such a distance metric for a type of input data, \eg, images or texts, that conforms to semantic distance measures between the data instances. 
It is typically achieved by learning an embedding function that maps similar instances to nearby points on a manifold in the embedding space and dissimilar instances apart from each other. 
Along with the recent advance in deep neural networks, deep metric learning has evolved and applied to a variety of tasks such as image retrieval \cite{schroff2015facenet}, person re-identification \cite{chen2017beyond} and visual tracking \cite{son2017multi}.
In contrast to conventional classification approaches, which learn category-specific concepts using explicit instance-level labels for predefined classes, 
metric learning learns the general concept of distance metrics using relational labels between samples in the form of pairs or triplets. 
This type of learning is natural for information retrieval, \eg, image search, where the goal is to return instances that are most similar to a query, and is also a powerful tool for open-set problems where we match or classify instances of totally new classes based on the learned metric.  
For this reason, metric learning has focused on generalization to unseen classes, that have never been observed during training~\cite{hoffer2015deep,schroff2015facenet,wohlhart2015learning}. 
Despite recent progress of deep metric learning, however, it remains exceedingly challenging for the learned embedding to generalize to unseen classes with a substantial domain gap.

\begin{figure}[t!]
    \centering
    \includegraphics[width=0.8\linewidth]{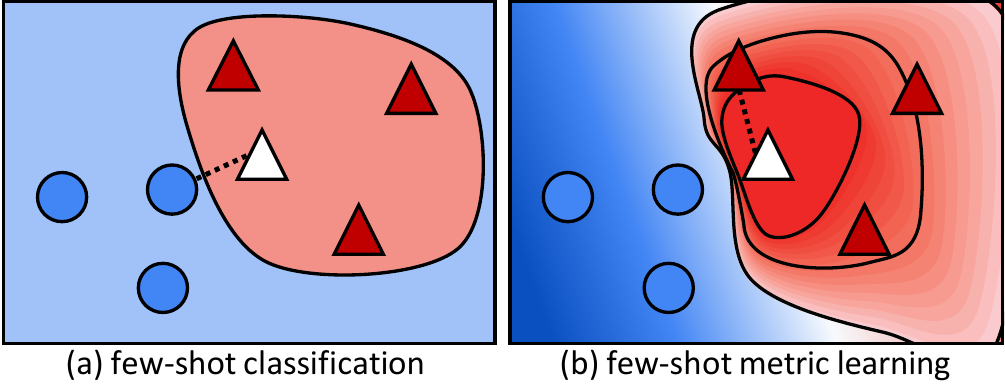}
    \caption{\textbf{Few-shot classification vs. few-shot metric learning.}
(a) A few-shot classifier learns to construct a decision boundary between support examples of different classes (red and blue). The query (white) is correctly classified as the red class, but fails to be placed away from blue samples.
(b) A \textit{few-shot metric learner} learns improved distance relations by adapting the embedding space online using the support examples, where the contour line represents the points with the same distance from the query.}
    \label{fig:teaser}
\end{figure}

To bridge the generalization gap in metric learning, we investigate the problem of few-shot metric learning that aims to {\em adapt an embedding function on the fly} to target classes with only a few annotated data. 
While the problem of few-shot learning has been actively studied for classification~\cite{koch2015siamese,ravi2016optimization}, the problem for metric learning has never been directly investigated so far to the best of our knowledge. 
Few-shot metric learning and classification problems share the goal of adapting to scarce labeled data, but diverge in terms of their training objectives and evaluation protocols, thus requiring different approaches.
As illustrated in Fig.~\ref{fig:teaser}, few-shot classification focuses on forming a decision boundary between samples of different classes and often fails to measure the relative distance relations between the samples, which are crucial for a retrieval task.
Our analysis also shows that the improvement of classification accuracy does not necessarily lead to that of retrieval accuracy (Table~\ref{table:comp_cls}).
While one of the main approaches to few-shot classification is to learn a metric space for nearest-neighbor classification to generalize to unseen classes~\cite{koch2015siamese,triantafillou2017few,vinyals2016matching,snell2017prototypical,ye2020few,feat,deepemd,li2021universal,chikontwe2022cad,renet}, it exploits the learned metric in testing without adapting the embedding space to the few-shot instances online. 
In this sense, metric-based few-shot classification is very different from few-shot metric learning that we investigate in this paper.

In this work we introduce three baselines for few-shot metric learning by adapting existing methods and propose the Channel-Rectifier Meta-Learning (\ours),  which effectively adapts the metric space online by adjusting channels of intermediate layers. 
We compare them to conventional metric learning as well as few-shot classification counterparts on \mini ImageNet~\cite{ravi2016optimization}, CUB-200-2011~\cite{wah2011caltech}, and MPII~\cite{kwak2016thin}.
We also introduce a new multi-attribute dataset for image retrieval, dubbed \mini DeepFashion, where two instances may have an opposite similarity relationship over multiple attributes (perspectives), and thus embedding functions are required to adapt to target attributes.
Experiments show that \ours~as well as the three baselines significantly improve image retrieval quality by adapting the model to target classes.
Notably, such improvement is significant when the gap between training and testing domains arises, which conventional metric learning often fails to bridge.

\section{Related work}
\subsection{Metric learning}
Metric learning has been studied extensively in the past decades~\cite{short1981optimal}, and has shown a great success recently using deep embedding networks trained with new losses.
One of the widely studied losses is the pair-wise  loss~\cite{chopra2005learning,schroff2015facenet,oh2016deep,wang2019multi} which minimizes the distance between two instances that have the same label and separates them otherwise.
Such losses include contrastive loss~\cite{bromley1994signature,chopra2005learning,hadsell2006dimensionality}, triplet loss~\cite{schroff2015facenet,wang2014learning}, lifted structured loss~\cite{oh2016deep}, and multi-similarity loss~\cite{wang2019multi}.
Unlike pair-based losses, proxy-based losses~\cite{movshovitz2017no,aziere2019ensemble,qian2019softtriple,kim2020proxy} associate proxy embeddings with each training class as a part of learnable parameters and learn semantic distance between an instance and the proxies.
These previous methods, which we refer to as {\em conventional} metric learning, emphasize the generalization performance on unseen classes that have never been observed during training.
However, they often suffer from a significant gap between source and target classes~\cite{milbich2021characterizing}. 
Although it is very practical to utilize a few labeled data from the target classes on the fly, online adaptation of the metric has never been explored so far to the best of our knowledge.
Recently, Milbich~\etal~\cite{milbich2021characterizing} showcase the effect of few-shot adaptation as a mean of out-of-distribution deep metric learning, their work does not present a problem formulation and a method dedicated to few-shot learning while our work does both of them.
In many practical applications of metric learning, a metric can also be learned with continuous labels of similarity , which are more informative but costly to annotate~\cite{sumer2017self,gordo2017beyond,kwak2016thin,kim2019deep}.
Online adaptation of metric learning may be particularly useful for such a scenario where we need to adapt the metric to the target classes with their few yet expensive labels.

\subsection{Few-shot classification}

Few-shot learning has been actively investigated for classification problems, and recent work related to ours is roughly categorized into three types: metric-based, optimization-based, and transfer-learning methods. 
The key idea of metric-based methods~\cite{koch2015siamese,triantafillou2017few,vinyals2016matching,snell2017prototypical,ye2020few,feat,deepemd,li2021universal,chikontwe2022cad,renet} is to learn an embedding space via episodic training so that the class membership of a query is determined based on its nearest class representations in the embedding space.
Although the metric-based few-shot classification and few-shot metric learning share the terminology ``metric'', they clearly differ from each other in terms of their learning objectives and practical aspects.
While metric-based few-shot classifiers construct a decision boundary using a learned metric without online adaptation of embeddings, few-shot metric learners learn improved metric function by adapting the embeddings online. 
In this aspect, $N$-way $1$-shot retrieval task, proposed in Triantafillou \textit{et al.}~\cite{triantafillou2017few}, is different from our task, few-shot metric learning. 
The $N$-way $1$-shot retrieval task in~\cite{triantafillou2017few} does not perform any online adaptation of embedding in the inference time; thus, it is exactly the same as the conventional deep metric learning. 
In this work, we focus on instance retrieval problems on an adaptive embedding space of a few examples, while class discrimination is out of our interest.

The optimization-based few-shot classification methods~\cite{ravi2016optimization,finn2017model,sun2019meta} learn how to learn a base-learner using a few annotated data.
While two aforementioned lines of work follow meta-learning frameworks, recent studies suggest that the standard transfer learning is a strong baseline for few-shot classification~\cite{chen2019closer,wang2020few,hu2022pushing,negmargin,chowdhury2021few}.
Such transfer learning methods pre-train a model using all available training classes and leverage the model for testing.

The contribution of this paper is four-fold:
\begin{itemize}
  \item We introduce a new problem of \textit{few-shot metric learning} that aims to adapt an embedding function to target classes with only a few annotated data. 
  
  \item We present three few-shot metric learning baselines and a new method, Channel-Rectifier Meta-Learning (CRML), which tackles the limitations of the baselines.
  
  \item We extensively evaluate them on standard benchmarks and demonstrate that the proposed methods outperform the conventional metric learning and few-shot classification approaches by a large margin.
  
   \item We introduce \mini DeepFashion, which is a challenging multi-attribute retrieval dataset for few-shot metric learning.
   
\end{itemize}


\section{Few-shot metric learning}

The goal of few-shot metric learning is to learn an embedding function for target classes with a limited number of labeled instances. In this section, we first revisit conventional metric learning, and then introduce the problem formulation and setup of few-shot metric learning.

\subsection{Metric learning revisited}

Let us assume data $\mathcal{X}$ of our interest, \eg, a collection of images. Given an instance $x \in \mathcal{X}$, we can sample its positive example $x^+$, which is from the same class with $x$, and its negative example $x^-$, which belongs to a different class from $x$. 
The task of metric learning is to learn a distance function $d$ such that 
\begin{equation}
\forall (x, x^+, x^-), \,\, d(x, x^+; \theta) < d(x, x^-; \theta). 
\end{equation}
Deep metric learning solves the problem by learning a deep embedding function $f(\cdot,\theta)$, parameterized by $\theta$, that projects instances to a space where the Euclidean distance is typically used as a distance function $d$:
\begin{equation}
d(x, x'; \theta) = \lVert f(x;\theta) - f(x';\theta) \rVert^2_2.
\end{equation} 
Note that metric learning focuses on unseen class generalization. 
The conventional setup for metric learning~\cite{oh2016deep} assumes a set of training classes $\mathcal{C}^\text{tr}$ and its dataset $\mathcal{D}^\text{tr}=\{ (x_t, y_t) | y_t \in \mathcal{C}^\text{tr} \}_{t}$ that contains labeled instances $x_t$ of the training classes. 
The task of metric learning then is to learn an embedding model using the dataset $\mathcal{D}^\text{tr}$ so that it generalizes to a dataset of \textit{unseen classes} $\mathcal{D}^\text{un} =\{(x_u, y_u) | y_u \in \mathcal{C}^\text{un} \}_{u}$, which contains instances from the classes not observed in training, \ie, $\mathcal{C}^\text{tr} \cap \mathcal{C}^\text{un} = \emptyset$.

\begin{figure*}[t!]
    \centering
    \includegraphics[width=\textwidth]{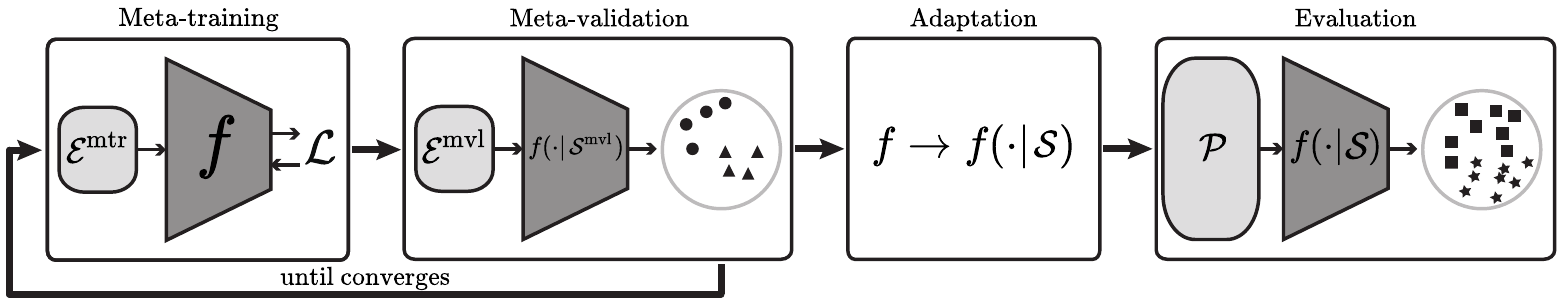}
    \caption{The problem formulation of few-shot metric learning as an episodic training. }
    \label{fig:overview}
\end{figure*}

\subsection{Problem formulation of few-shot metric learning}

In contrast to the conventional metric learning, few-shot metric learning uses annotated instances of the target classes (thus, not totally unseen any more) but only a few instances per class, which is reasonable in most real-world scenarios. 
For simplicity and fair comparison, we applied $N$-way $K$-shot setting. 
Let us assume a target class set $\mathcal{C}$, which contains $N$ classes of our interest, and its \textit{support set} $\mathcal{S}=\{ (x_s, y_s) | y_s \in \mathcal{C} \}^{NK}_{s=1}$, which contains $K$ labeled instances for each of the target classes;
$K$ is supposed to be small. 
The task of $N$-way $K$-shot metric learning is to learn an embedding model using the support set $\mathcal{S}$ so that it generalizes to a \textit{prediction set} $\mathcal{P}=\{(x_p, y_p) | y_p \in \mathcal{C} \}^{M}_{p=1}$, which contains $M$ unseen instances, \ie, $\mathcal{S} \cap \mathcal{P} = \emptyset$, from the target classes. 
The generalization performance is evaluated using instance retrieval on $\mathcal{P}$.

Our goal is to learn such a few-shot metric learning model using a set of labeled instances from {\em non-target} classes, which can be viewed as a meta-learning problem. 
In training a few-shot metric learning model, we thus adopt the episodic training setup of meta-learning~\cite{vinyals2016matching,ravi2016optimization,finn2017model,snell2017prototypical} as illustrated in Fig. \ref{fig:overview}. 
In this setup, we are given a meta-train set $\mathcal{D}^\text{mtr}$ and a meta-validation set $\mathcal{D}^\text{mvl}$. 
They both consist of labeled instances from non-target classes but their class sets, $\mathcal{C}^\text{mtr}$ and $\mathcal{C}^\text{mvl}$, are disjoint, \ie, $\mathcal{C} \cap \mathcal{C}^\text{mtr} =  \mathcal{C} \cap \mathcal{C}^\text{mvl} = \mathcal{C}^\text{mtr} \cap \mathcal{C}^\text{mvl} = \emptyset$.

A meta-train episode is constructed from $\mathcal{D}^\text{mtr}$ by simulating a support set and its prediction set; (1) a support set $\mathcal{S}^\text{mtr}$ is simulated by sampling $N$ classes from $\mathcal{C}^\text{mtr}$ then $K$ instances for each of the $N$ classes, and (2) a prediction set $\mathcal{P}^\text{mtr}$, which is disjoint from $\mathcal{S}^\text{mtr}$, is simulated by sampling other $K'$ instances for each of the $N$ classes. 
A meta-validation episode is constructed from $\mathcal{D}^\text{mvl}$ likewise. 

The meta-trained few-shot metric learning model is tested on meta-test set $\mathcal{E} = \{ (\mathcal{S}_m, \mathcal{P}_m) \}_m$ from the target classes $\mathcal{C}$.

\section{Methods}

We introduce three baselines for few-shot metric learning (Sec.~\ref{sec:sft}-\ref{sec:mtl}) by adapting existing few-shot learning methods, representative for few-shot classification and appropriate for few-shot metric learning in the sense that it adapts the embedding spaces online.
We then discuss the limitations of the baselines and propose our method that overcomes the limitations (Sec.~\ref{sec:crml}). \
\subsection{Simple Fine-Tuning (SFT)}
\label{sec:sft}
As the first simple baseline for few-shot metric learning, we use the standard procedure of inductive transfer learning. 
In training, the embedding model $f$ is trained on meta-training set $\mathcal{D}^\text{mtr}$ from scratch with a metric learning loss; this is done in the same way with conventional metric learning, not involving any episodic training.
In testing with an episode $(\mathcal{S}, \mathcal{P})$, the trained model $f(x;\theta_0)$ is simply fine-tuned using the target support set $\mathcal{S}$ by computing gradients with respect to a metric learning loss on $\mathcal{S}$: 
\begin{equation}
    \theta' = \theta_0 - \alpha \nabla_\theta \mathcal{L}(S;\theta_0 ).
\end{equation}
After fine-tuning, the model is tested on $\mathcal{P}$.
We choose the number of updates that shows the highest performance on the (meta-)validation set.

\subsection{Model-Agnostic Meta-Learning (MAML)} 
\label{sec:maml}
As the second baseline for few-shot metric learning, we employ MAML~\cite{finn2017model}, which meta-learns a good initialization for few-shot adaptation via episodic training.
Given a meta-train episode $(\mathcal{S}_k^\text{mtr}, \mathcal{P}_k^\text{mtr})$, the meta-training process consists of inner and outer loops. 
In the inner loop, the parameters of a {\em base-learner} are updated using a {\em meta-learner} and the support set $\mathcal{S}_k^\text{mtr}$. 
In MAML, the base-learner corresponds to the embedding model $f$, and the meta-learner corresponds to the initializer $\theta_0$ for the embedding model.
The inner loop updates the base learner using $\mathcal{S}_k^\text{mtr}$ by a few gradient descent steps:
\begin{equation}
\theta_1 = \theta_0 - \alpha \nabla_\theta \mathcal{L}(\mathcal{S}^{\text{mtr}}_k;\theta_0),
\label{eq:maml_inner}
\end{equation} 
where $\alpha$ is the step size of inner-loop updates.
In the outer loop, the meta-learner is updated using the loss of the updated base-learner  on the prediction set $\mathcal{P}_k^{\text{mtr}}$:
\begin{equation}
    \theta_0' = \theta_0 - \eta \nabla_\theta {\mathcal{L}(\mathcal{P}^{\text{mtr}}_k;\theta_1)},
    \label{eq:maml_outer}
\end{equation}
where $\eta$ is the step size of outer-loop updates. 
This meta-optimization with episodic training seeks to learn the initialization of the embedding model that generalizes well to unseen classes. 
For computational efficiency, we use the first-order approximation of meta-gradient update as in~\cite{finn2017model}. 

The meta-testing procedure with an episode $(\mathcal{S}, \mathcal{P})$ is the same as that of the inner loop; the meta-trained model $f(x;\theta_0')$ is evaluated by fine-tuning it on $\mathcal{S}$ with a few gradient steps and testing it on $\mathcal{P}$.
We choose the number of steps that shows the highest performance on the $\mathcal{D}^\text{mvl}$.

\subsection{Meta-Transfer Learning (MTL)}
\label{sec:mtl}
As the third baseline, we adapt MTL~\cite{sun2019meta} for few-shot metric learning. 
MTL first pre-trains an embedding model $f$ on the meta-training set $\mathcal{D}^{\text{mtr}}$ and then conducts an optimization-based meta-learning process while freezing all the layers except for the last fully-connected one.
Here, the base-learner is the last fully-connected layer $\psi$ and 
the meta-learner consists of two groups: a set of channel scaling/shifting parameters $\Phi=\{(\gamma^l,\beta^l)\}_l$ and the initialization of the last fully-connected layer $\psi_0$. 
The channel-scaling/shifting parameters are applied to the frozen convolutional layers of the embedding model $f$ 
so that the parameters of each conv layer are scaled by $\gamma^l$ and shifted by $\beta^l$: 
\begin{equation}
    \mathrm{Conv}_\text{CSS}^l(X;W^l,b^l,\gamma^l,\beta^l)=(W^l\odot \gamma^l)\ast X +(b^l+\beta^l), 
    \label{eq:mtl}
\end{equation}
where $W^l$ and $b^l$ are the weight and bias of each convolution layer, $\odot$ is channel-wise multiplication, and $\ast$ is convolution. 
For example, if the $3\times3$ convolution kernel is size of $128\times64\times3\times3$, then channel-scaling parameter is size of $128\times64\times1\times1$. 

In the inner loop of the meta-training, the last fully-connected layer is fine-tuned on $\mathcal{S}_k^{\text{mtr}}$ from the initialization $\psi_0$. 
In the outer loop,  the set of channel-scaling/shifting parameters $\Phi$ and the initialization of the last fully-connected layer $\psi_0$ are meta-updated using the prediction loss. 

In meta-testing with an episode $(\mathcal{S}, \mathcal{P})$, following the process of the inner loop, only the last layer $\psi$ is updated via a few gradient steps from the meta-learned initialization $\psi_0$ using $\mathcal{S}$, and the fine-tuned model is tested on $\mathcal{P}$.

\subsection{Channel-Rectifier Meta-Learning (CRML)}
\label{sec:crml}
In this subsection we discuss limitations of the aforementioned baselines and then propose a simple yet effective method for few-shot metric learning.

The main challenge in few-shot metric learning is how to effectively adapt the vast metric space using only a few examples while avoiding the danger of overfitting; as expected, the issue of overfitting is particularly critical in few-shot learning since only a few annotated examples are given for adaptation online.  
Updating all learnable parameters during meta-testing, as typically done in simple fine-tuning (Sec.~\ref{sec:sft}) and MAML (Sec.~\ref{sec:maml}), often causes quick over-fitting. 
A reasonable alternative is to fine-tune a part of the network only,  \eg, the output layer as in MTL (Sec.~\ref{sec:mtl}), with all the other parts frozen.
This partial update strategy is shown to be effective for classification problems where class decision boundaries can be easily affected by the specific part.  
In metric learning, however, fine-tuning the output layer or other specific layer online turns to hardly change the metric space, \ie, distance relations among embeddings (Sec.~\ref{sec:exp_fsml}). 

To tackle the issue, we propose the \emph{Channel-Rectifier Meta-Learning} (\ours) that meta-learns {\em how to rectify} channels of intermediate feature layers.  
The main idea is inspired by channel scaling/shifting of MTL~(Eq.~(\ref{eq:mtl})), which is shown to be effective in adapting pretrained layers but never used in online adaptation.   
Unlike MTL, we propose to leverage the channel scaling/shifting module, dubbed {\em channel rectifier}, for online adaptation.
In other words, we set the channel rectifier $\Phi=\{(\gamma^l,\beta^l)\}_l$ as a base-learner and its initialization $\Phi_0$ as a meta-learner. 
In this setup, we pre-train an embedding model $f$ on the meta-training set $\mathcal{D}^{\text{mtr}}$ and all the pre-trained parameters are frozen during the subsequent meta-learning process.
Instead, in the meta-training stage, we update the channel rectifier $\Phi$ in the inner loop~(Eq.~(\ref{eq:maml_inner})) while updating the initialization of the channel rectifier $\Phi_0$ in the outer loop.  
This meta-learning process of CRML is summarized in Alg. \ref{alg:crml}, where we describe a single-step inner loop with meta-batch of size 1 for the sake of simplicity.

In meta-testing with an episode $(\mathcal{S}, \mathcal{P})$, only the channel rectifier $\Phi$ is fine-tuned by the support set $\mathcal{S}$ with a few gradient steps from the learned initialization $\Phi_0$, and the fine-tuned model is tested on $\mathcal{P}$.
Note that the \ours allows the channel rectifier to effectively exploit the support set to adapt the embedding function online.

\setlength{\textfloatsep}{10pt}
\begin{algorithm}[t!]
\begin{minipage}[t][][b]{\textwidth}
\caption{Channel-Rectifier Meta-Learning}
\label{alg:crml}
\textbf{Input}: Meta training set $\mathcal{D}^\text{mtr}$, learning rate $\alpha, \eta$, pre-trained embedding model $f$ \\
\textbf{Output}: The set of initialization of channel-scaling/shifting parameter $\Phi_0=\{(\gamma_0^l,\beta_0^l)\}_l$
\begin{algorithmic}[1]
\STATE Initialize $\gamma^l \leftarrow \mathbf{1}, \beta^l \leftarrow \mathbf{0}$ \;
\FOR{$(\mathcal{S}_k^{\text{mtr}}, \mathcal{P}^{\text{mtr}}_k) \in D^\text{mtr}$}
\STATE $\Phi_1 \leftarrow \Phi_0 - \alpha\nabla_\Phi \mathcal{L}(\mathcal{S}_k^{\text{mtr}}; f, \Phi_0) $ // Inner loop  \;
\STATE $\Phi_0 \leftarrow \Phi_0 - \eta \nabla_\Phi \mathcal{L}(\mathcal{P}_k^{\text{mtr}}; f, \Phi_1)$ // Outer loop \;
\ENDFOR
\end{algorithmic}
\end{minipage}
\end{algorithm}

\setlength{\textfloatsep}{20.0pt plus 2.0pt minus 4.0pt}

\section{Experiments}

\subsection{Experimental Settings}
\label{sec:datasets}

\subsubsection{Datasets and scenarios.} 
We evaluate CRML and three baselines on two standard few-shot learning datasets, \textit{mini}ImageNet~\cite{ravi2016optimization} and CUB-200-2011~\cite{wah2011caltech}. 
The \textit{mini}ImageNet dataset is a subset of the ImageNet~\cite{krizhevsky2012imagenet} and consists of 60,000 images categorized into 100 classes with 600 images each. 
We use the splits divided into 64/16/20 classes for (meta-)training, (meta-)validation, and (meta-)testing, which has been introduced by~\cite{ravi2016optimization}. 
The CUB-200-2011 (CUB) is a fine-grained classification dataset of bird species. 
It consists of 200 classes with 11,788 images in total. 
Following the evaluation protocol of \cite{hilliard2018few}, the split is divided into 100/50/50 species for (meta-)training, (meta-)validation, and (meta-)testing. 
We also use the MPII dataset, a human pose dataset, to train the metric learning model with continuous labels. 
Following the split of~\cite{kwak2016thin}, 22,285 full-body pose images are divided into 12,366 for (meta-)training and 9,919 for (meta-)testing. 
Following~\cite{kim2019deep}, the label for a pose pair is defined as a pose distance, the sum of Euclidean distances between body-joint locations. 
To verify the influence of the domain gap between the source and the target classes, we conduct cross-domain experiments~\cite{chen2019closer}, which is designed to have a significant domain gap between training and evaluation; (meta-)training set consists of all samples in \mini ImageNet where each (meta-)validation and (meta-)test set consists of 50 classes from CUB.  
Lastly, we propose a new multi-attribute dataset for few-shot metric learning, dubbed \textit{mini}DeepFashion. 
The \mini DeepFashion is built on DeepFashion~\cite{liuLQWTcvpr16DeepFashion} dataset, which is a multi-label classification dataset with six fashion attributes. 
It consists of 491 classes and the number of all instances in the dataset amounts to 33,841. 
The details of \mini DeepFashion is in  Sec.~\ref{sec:results_fashion}.

\label{sec:evaluation_metrics}
\subsubsection{Evaluation metrics.}
We use two standard evaluation metrics, mAP (mean value of average precision)~\cite{boyd2013area} and Recall$@k$~\cite{jegou2010product}, to measure image retrieval performances. 
Recall$@k$ evaluates the retrieval quality beyond $k$ nearest neighbors while mAP evaluates the full ranking in retrieval. 
Since the MPII dataset for the human pose retrieval is labeled with continuous real values, we employee two metrics defined on continuous labels following~\cite{kim2019deep}: mean pose distance (mPD) and a modified version of normalized discounted cumulative gain (nDCG). 
The $\textrm{mPD}_k$ evaluates the mean pose distance between a query and $k$ nearest images. 
The modified $\textrm{nDCG}_k$ evaluates the rank of the $k$ nearest images and their relevance scores. 
The details about evaluation metrics are specified in the supplementary material.

\subsubsection{Implementation details.}
We use ResNet-18~\cite{he2016deep} for the main backbone from scratch.
We append a fully-connected layer with the embedding size of 128 followed by \textit{l}2 normalization on top of the backbone. 
We use the multi-similarity loss~\cite{wang2019multi} for training all the baselines and ours, CRML.
For the human pose retrieval task on the MPII, we use ResNet-34 with the embedding size of 128, which is pre-trained on ImageNet~\cite{krizhevsky2012imagenet} for a fair comparison with~\cite{kim2019deep}. 
We fine-tune the network with log-ratio loss~\cite{kim2019deep} using from 25 to 300 pairs out of all possible ${12366 \choose 2} \approx 7.6 \times 10^7$ pairs. 
Complete implementation details are specified in the supplementary material.

\begin{table*}[t!]
\begin{minipage}[t][][b]{0.49\textwidth}
\centering \fontsize{8}{10}\selectfont
\setlength{\tabcolsep}{1pt}
\caption{Performance on \textit{mini}ImageNet.}
\scalebox{0.80} {
\begin{tabular}{lccccccccc}
   & \multicolumn{4}{c}{5-way 5-shot} & & \multicolumn{4}{c}{20-way 5-shot}  \\ 
Method      & mAP & R@1 & R@2 & R@4 & & mAP & R@1 & R@2 & R@4  \\ 
\hline
\hline
DML         &46.9 &73.5 &84.1 &91.2 & &21.0 &49.2 &62.6 &74.2  \\ \hline
$\text{FSML}_{\text{SFT}}$   &63.3 &78.6 &85.7 &91.8 & &29.9 &52.6 &65.1 &75.3    \\  
$\text{FSML}_{\text{MAML}}$   &65.9 &79.8 &87.5 &92.2 & &28.6 &52.7 &65.6 &76.3  \\ 
$\text{FSML}_{\text{MTL}}$   &56.5 &77.2 &86.7 &92.6 & &24.0 &50.0 &64.6 &76.2  \\ 
\ours & \textbf{69.2} & \textbf{83.2}  & \textbf{89.9} & \textbf{93.8} & & \textbf{30.7} & \textbf{56.3} & \textbf{68.6} & \textbf{78.5} \\
\hline
\end{tabular} 
}
\label{table:many_way_mini}
\end{minipage}
\begin{minipage}[t][][b]{0.49\textwidth}
\centering \fontsize{8}{10}\selectfont

\setlength{\tabcolsep}{1pt}
\caption{Performance on CUB-200-2011.}
\scalebox{0.80} {
\begin{tabular}{lccccccccc}
 & \multicolumn{4}{c}{5-way 5-shot} & & \multicolumn{4}{c}{50-way 5-shot}  \\ 
Method      & mAP & R@1 & R@2 & R@4  && mAP & R@1 & R@2 & R@4  \\
\hline
\hline
DML         &57.8 &81.4 &88.6 &93.6  &&26.3 &51.8 &62.5 &72.2 \\  \hline
$\text{FSML}_{\text{SFT}}$   &79.9 &87.7 &91.6 &93.8  && 31.1 &55.3 &66.2 &74.8  \\
$\text{FSML}_{\text{MAML}}$   &82.0 &89.5 &93.2 &95.4  &&33.2 &54.6 &66.3 &75.7  \\ 
$\text{FSML}_{\text{MTL}}$   &71.9 &86.2 &91.4 &94.5  &&30.2 &54.3 &65.2 &73.6  \\ 
\ours & \textbf{82.7} & \textbf{90.0}  & \textbf{93.5} & \textbf{95.5} & & \textbf{33.9} & \textbf{58.1} & \textbf{68.4} & \textbf{76.5} \\
\hline
\end{tabular}  
}
\label{table:many_way_cub} 
\end{minipage}
\end{table*}

\begin{table*}[t!]
\begin{minipage}[t][][b]{0.49\textwidth}
\centering\fontsize{8}{10}\selectfont
\setlength{\tabcolsep}{1pt}

\caption{Performance on cross-domain.\label{table:cross_domain}}
\scalebox{0.80} {
\begin{tabular}{lccccccccccc} 
 & \multicolumn{4}{c}{5-way 5-shot} & & \multicolumn{4}{c}{50-way 5-shot}  \\
Method          & mAP & R@1 & R@2 & R@4  && mAP & R@1 & R@2 & R@4  \\
\hline
\hline
DML             &36.2  &57.5 &73.2 &86.1  & &6.1  &19.6 &29.0 &41.1  \\ \hline
$\text{FSML}_{\text{SFT}}$       &49.4  &65.2 &77.1 &86.0  & &9.8  &24.1 &35.2 &47.8  \\ 
$\text{FSML}_{\text{MAML}}$ &51.5  &67.0 &78.8 &87.2  & &10.0 &23.7 &35.4 &48.8 \\ 
$\text{FSML}_{\text{MTL}}$ &40.3  &62.7 &28.7 &41.1  & &6.8 &20.0 &29.9 &42.8 \\ 
CRML & \textbf{56.4} & \textbf{71.0} & \textbf{81.5} & \textbf{88.9} & & \textbf{10.9} & \textbf{27.0} & \textbf{38.3} & \textbf{51.1} \\
\hline
\end{tabular} 
}
\end{minipage}
\hfill
\begin{minipage}[t][][b]{0.49\textwidth}
\centering\fontsize{8}{10}\selectfont
\setlength{\tabcolsep}{1pt}
\caption{Performance on \textit{mini}DeepFashion.} \label{table:deep_fashion}
\scalebox{0.80} {
\begin{tabular}{lccccccccc} 
 & \multicolumn{4}{c}{5-way 5-shot} & & \multicolumn{4}{c}{20-way 5-shot}  \\
Method          & mAP & R@1 & R@2 & R@4 & &mAP & R@1 & R@2 & R@4  \\ 
\hline
\hline
DML             &31.8 &50.3 &65.8 &80.2  & &11.3 &26.1 &37.3 &50.2  \\ \hline
$\text{FSML}_{\text{SFT}}$       &35.2 &51.3 &66.1 &79.8 & &12.5 &26.4 &37.8 &50.4  \\ 
$\text{FSML}_{\text{MAML}}$ &38.2 &\textbf{53.5 }&\textbf{67.6} &80.2 & &\textbf{13.3} &27.7 &\textbf{39.1} &\textbf{52.1}   \\  
$\text{FSML}_{\text{MTL}}$ &35.2 &52.2 &66.8 &79.9 & &12.3 &26.8 &38.0 &50.6   \\  
CRML & \textbf{38.3} & 50.7 & 66.3 & \textbf{80.2} & & 13.0 & \textbf{27.8} & \textbf{39.1} & 52.0 \\
\hline
\end{tabular}
}
\end{minipage}
\end{table*}

\subsection{Effectiveness of few-shot metric learning}
\label{sec:exp_fsml}
\subsubsection{Few-shot metric learning is effective on discrete-label benchmarks.}
We first compare few-shot metric learning to conventional metric learning (DML) in Tables \ref{table:many_way_mini}, \ref{table:many_way_cub}, and \ref{table:cross_domain}. 
All the few-shot metric learning methods consistently outperform DML not only on the 5-way 5-shot setting, which is standard for few-shot learning but also on the full-way 5-shot setting, which is standard for image retrieval.
The result also shows that only five shots for each class is enough to boost the image retrieval quality regardless of the number of classes to retrieve.
Such improvement is clear not only in Recall@$k$, \ie, the measurement of top-$k$ nearest neighbors, but also in mAP, \ie, the quality of all distance ranks.
More importantly, the proposed CRML outperforms all baselines in the most settings, improving over MAML and SFT baselines by a large margin on the \mini ImageNet and the cross domain setting.
CRML is trained to \textit{rectify} the base feature maps by learning a small set of channel scaling and shifting parameters and thus effectively avoids overfitting to the few-shot support set from an unseen domain.
In contrast, both the MAML and the SFT baselines update all parameters in the embedding functions online, thus being vulnerable to overfitting to the small number of support set.
Note that the worst model is the MTL baseline, which fine-tunes the last fully-connected layer while all the other layers frozen, suggesting that fine-tuning only a single output layer in the embedding function is insufficient for online adaptation.
Note that simple fine-tuning (SFT) often performs comparable to meta-learning baselines (MAML and MTL) as recently reported in transfer learning based few-shot learning work~\cite{chen2019closer}.

\subsubsection{Few-shot metric learning is effective on continuous-label benchmarks.}
We also evaluate few-shot metric learning on the human pose retrieval on MPII to demonstrate its applications and show its effectiveness. 
Given a human image with a certain pose, the goal of the human pose retrieval is to retrieve the most similar image of a human pose, where the supervisions in MPII consist of a continuous value on a pair-wise similarity. 
Since such labels are expensive to collect, few-shot learning is a practical solution for this problem, while the standard image classification approach is unlikely to applied due to the pair-based form of supervisions.
Table~\ref{table:human_pose} summarizes the retrieval performances with increasing numbers of the pair-wise supervisions. 
The retrieval performance gradually improves as the number of labels increase, although the source classes (object classes in ImageNet~\cite{krizhevsky2012imagenet}) used for pre-training deviate from the target classes of human poses. 
Few-shot adaptation achieves 7.7\% of the state-of-the-art performance~\cite{kim2019deep} trained with full supervisions using only 0.00039\% of supervisions. 

Please refer to the supplementary material for qualitative visualizations of our method and more experiments about effectiveness of few-shot metric learning with 1) 10 shots, 2) additional metric learning losses, 3) qualitative results.

\begin{table*}[t!]
\begin{minipage}[t][][b]{0.49\textwidth}
\centering\fontsize{8}{10}\selectfont
\setlength{\tabcolsep}{1.5pt}
\caption{Performance on MPII~\cite{andriluka14cvpr}. For nDCG${}_1$, the higher the better, and for mPD${}_1$, the lower the better. 
$\dagger$ and $\ddagger$ denotes the performances from ~\cite{kwak2016thin} and \cite{kim2019deep}.
}
\scalebox{0.9} {
\begin{tabular}{lcccccccc} 
& \multicolumn{8}{c}{\#(pair labels used)} \\
Metric & 0 & 25 & 50 & 100 & 200 & 300 & 76M\textsuperscript{$\dagger$} & 76M\textsuperscript{$\ddagger$}\\
\hline
\hline
nDCG${}_1$ & 40.4& 41.2 & 41.5 & 41.9 & 42.4 & 43.0 & 70.8 & 74.2 \\
mPD${}_1$ & 31.5& 30.9 & 30.8 & 30.7 & 30.3 & 29.9 & 17.5 & 16.5 \\
\hline
\end{tabular} 
\label{table:human_pose}
}
\end{minipage}
\begin{minipage}[t][][b]{0.49\textwidth}
\centering\fontsize{8}{10}\selectfont
\setlength{\tabcolsep}{1pt}
\caption{Adaptation growth rate\protect\footnotemark (\%) of \ours on three datasets.
The models are trained and evaluated in the 5-way 5-shot setting on each dataset.}
\scalebox{0.9} {
\begin{tabular}{lccccccc} 
Metric $\rightarrow$ & \multicolumn{3}{c}{mAP (\%)} & & \multicolumn{3}{c}{recall@1 (\%)}  \\
Dataset $\rightarrow$  & CUB & mini & cross & & CUB & mini & cross   \\ 
\hline
\hline
before adapt.       & 72.2 & 58.1 & 41.8 & & 87.5 & 80.7 & 65.3  \\
after adapt. & 82.7 & 69.2 & 56.4 & & 90.0 & 83.2 & 71.0  \\ \hline
growth rate (\%)    & 14.5 & 19.0 & \textbf{34.8} & & 2.8 & 3.1 & \textbf{8.7}  \\
\hline
\end{tabular} \label{table:growth_rate}
}
\end{minipage} 
\footnotetext[1]{The growth rates are computed using the original raw values and then rounded up.}
\end{table*}

\subsection{Influence of domain gap between source and target}
To verify the influence of the domain gap between the source and the target classes, we conduct cross-domain experiments. 
Table~\ref{table:cross_domain} shows the results of the 5-way and 50-way 5-shot experiments on the cross-domain setting. 
Due to the substantial domain gap between the source and target classes, the performances are much lower than those on CUB experiment in Table~\ref{table:many_way_cub}. 
However, the performance improvement is between 1.5 and 2 times higher than that of CUB.
We observe that CRML results in remarkable improvement, which implies CRML is learned to properly rectify the base features adapted to given a support set.

We investigate the correlation between domain gap and effects of few-shot adaptation.
For \textit{mini}ImageNet, we randomly sample 60 instances from each target class in to match the prediction set size equal to that of CUB for a fair comparison.
Note that the CUB is fine-grained thus has the smallest domain gap, while the cross-domain setting has the biggest. 
For each dataset, we measure the ratio of performance improvement from online adaptation and refer to it as adaptation growth rate.
As shown in Table~\ref{table:growth_rate}, the growth rate increases as the domain gap arises.
It implies that few-shot metric learning is more effective when the target classes diverge more from the source classes.

\begin{table}[t!]
\centering\fontsize{9}{11}\selectfont
\caption{Classification and image retrieval performances of few-shot classification and few-shot metric learning methods on \mini ImageNet in a 5-way 5-shot setting. TTA stands for test-time adaptation via gradient descents.}
\begin{tabular}{lccc}
& \small{Classification} & \multicolumn{2}{c}{Image retrieval} \\
Method & Accuracy & Recall@1 & mAP \\ 
\hline 
\hline
\multicolumn{4}{c}{Transfer-based few-shot classification} \\
\hline
Baseline~\cite{chen2019closer} & 62.21 & 74.04 & 53.14 \\
Baseline++~\cite{chen2019closer} & 74.88 & 78.97 & 65.22 \\
\hline
\multicolumn{4}{c}{Metric-based few-shot classification} \\
\hline
MatchingNet~\cite{vinyals2016matching} & 67.42 & 69.70 & 51.10  \\
+TTA & 71.50 & 71.05 & 52.51  \\
ProtoNet~\cite{snell2017prototypical} & 74.46 & 71.04 & 51.73  \\
+TTA & 77.15 & 71.89 & 51.78  \\
FEAT~\cite{ye2020few} & 80.37 & 79.15 & 49.73 \\
+TTA & \textbf{80.59} & 79.31 & 51.72 \\
\hline
\multicolumn{4}{c}{Optimization-based few-shot classification} \\
\hline
MAML~\cite{finn2017model} & 68.80 & 75.81 & 57.03 \\
\hline
\multicolumn{4}{c}{Few-shot metric learning} \\
\hline
$\text{FSML}_{\text{SFT}}$ & 69.92 & 79.14 & 65.22 \\
$\text{FSML}_{\text{MAML}}$ & 72.69 & 79.77 & 65.86 \\
$\text{FSML}_{\text{MTL}}$ & 70.34 & 77.19 & 56.50 \\
\ours & 76.64 & \textbf{83.22} & \textbf{69.15} \\
\hline
\end{tabular}
\label{table:comp_cls}
\end{table}

\subsection{Few-shot metric learning vs. few-shot classification}

To verify the differences between few-shot metric learning and few-shot classification, we compare them both in image retrieval and classification. 
We evaluate different types of few-shot classification methods for comparison: transfer-based (Baseline~\cite{chen2019closer}, Baseline++~\cite{chen2019closer}), optimization-based (MAML~\cite{finn2017model}), and metric-based (MatchingNet~\cite{vinyals2016matching}, ProtoNet~\cite{snell2017prototypical}, FEAT~\cite{ye2020few}) methods.
Note that as mentioned earlier, unlike the transfer-based and optimization-based methods, the metric-based ones in their original forms do not use online adaptation to the given support on test time. For a fair comparison, we thus perform add-on online adaptation, which is denoted by TTA, for the metric-based methods by a few steps of gradient descent using the support set.

\textit{We observe that there is little correlation between the classification accuracy and the image retrieval performances} as shown in Table~\ref{table:comp_cls}. 
All the few-shot metric learning methods outperform few-shot classification methods on image retrieval in terms of Recall@1 and mAP, while their classification accuracies are lower than those of few-shot classification methods. 
Interestingly, only Baseline++ shows competitive results with few-shot metric learning on image retrieval; we believe it is because the learned vectors in Baseline++ behave similarly to proxies in proxy-based metric learning methods.  
The results imply that few-shot classification learning and few-shot metric learning are distinct and result in different effects indeed.
Note that even metric-based few-shot classification is not adequate for organizing the overall metric space, and additional test-time adaptation contributes insignificantly to improving the image retrieval performances.

\begin{table*}[t!]
\centering
\begin{minipage}[c][][b]{0.35\textwidth}
\centering
\caption{The split of \textit{mini}Deep Fashion.}
\begin{tabular}{ccc}
\multirow{2}{*}{splits}   & attribute  & \multirow{2}{*}{\#instances} \\ 
                    & (\#classes)  & \\ 
                    \hline \hline
\multirow{3}{*}{$\mathcal{C}^{\text{mtr}}$}  & fabric (99)  & \multirow{3}{*}{27,079}\\ 
                        & part (95)  &\\ 
                        & style (127) & \\ \hline
$\mathcal{C}^{\text{mvl}}$ & shape (76) & 7,600 \\ \hline
\multirow{2}{*}{$\mathcal{C}$}   & category (37) &  \multirow{2}{*}{8,685} \\
                                 & texture (57)  &  \\ 
                        \hline
\multicolumn{2}{c}{total} & 33,841 \\
\hline
\end{tabular}
\label{table:split_deep}
\end{minipage}
\hspace{2.0mm}
\begin{minipage}[c][][b]{0.6\textwidth}
\centering
\includegraphics[width=\linewidth]{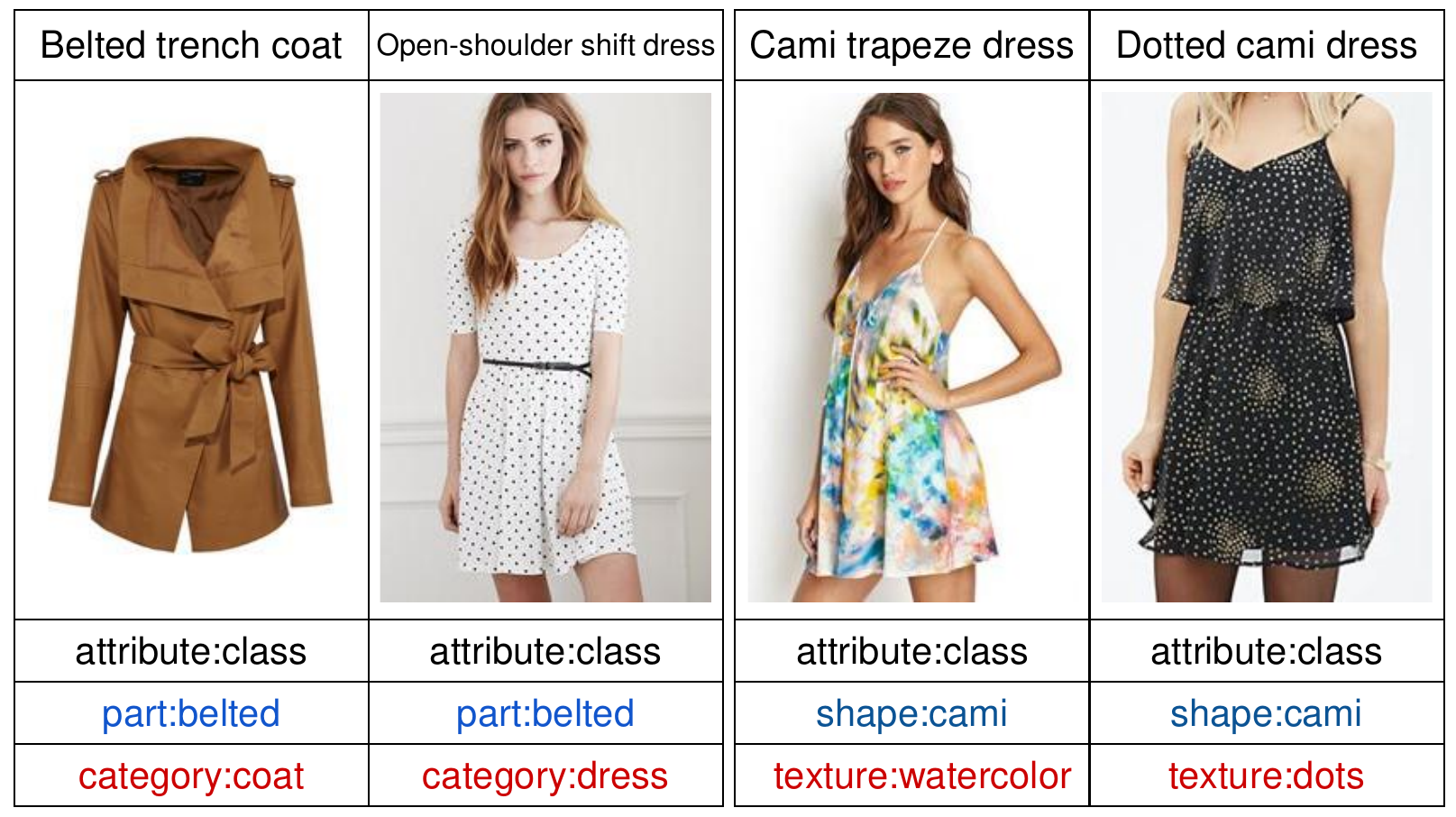}
\captionof{figure}{Two example pairs sharing attributes.
Blue rows denote positive pairs on one attribute, and red rows denote negative pairs on the other. 
}
\label{fig:fashion}
\end{minipage}
\end{table*}

\subsection{Results on \textbf{\mini}DeepFashion}
\label{sec:results_fashion}
Our design principle for the \mini DeepFashion is to make a binary (similar/dissimilar) relationship between two instances inconsistent across the non-target and target classes.
We thus split six attributes, each of which indicates a semantic perspective that categorizes instance into (meta-)training, (meta-)validation, and (meta-)testing as shown in Table~\ref{table:split_deep}. 
For example, a trench coat and a shift dress are either a positive pair in terms of parts they share, or a negative pair in terms of their categories as shown in Fig.~\ref{fig:fashion}.
As existing few-shot learning datasets~\cite{ravi2016optimization,wah2011caltech} do not assume such semantic switch, one embedding space is enough for a global information.
In contrast, the assumption is no longer valid on \mini DeepFashion, thus online adaptation is inevitable;
this feature is well-aligned with the goal of few-shot metric learning.
\mini DeepFashion is built on DeepFashion~\cite{liuLQWTcvpr16DeepFashion} dataset, which is a multi-label classification dataset with six fashion attributes. 
We construct \textit{mini}DeepFashion by randomly sampling 100 instances from each attribute class. 
The number of classes in each attribute and the number of instances in each split is shown in Table~\ref{table:split_deep}. Also, the class configuration for each attribute is in the supplementary materials. 

\begin{figure}[t!]
    \centering
    \includegraphics[width=\linewidth]{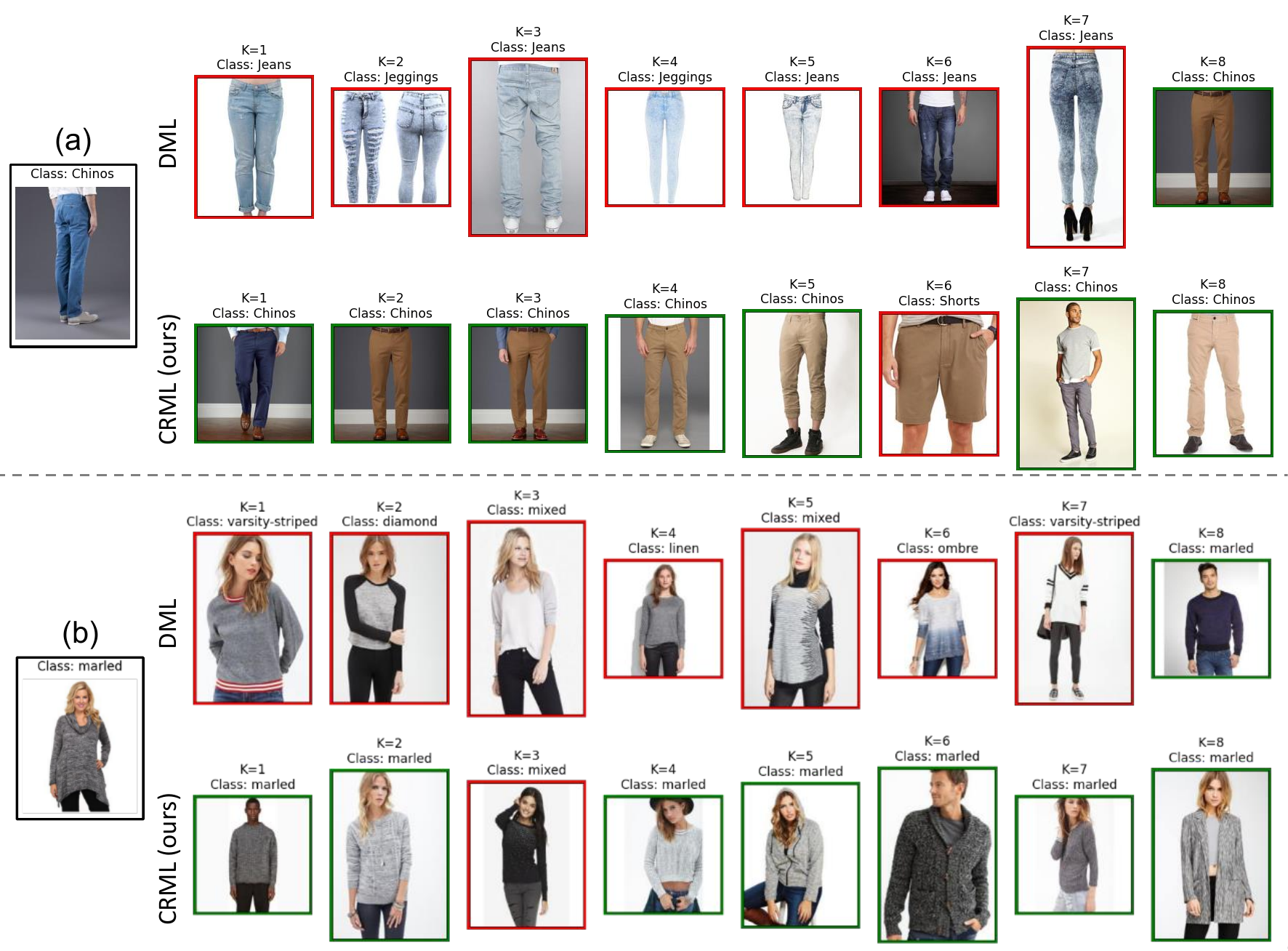}
    \caption{Retrieval results on the (a) texture and (b) category attribute in \textit{mini}- DeepFashion. The leftmost images are queries, and the right images are top-eight nearest neighbors. Green and red boxes are positive and negative images.}
    \label{fig:qual_fashion}
\end{figure}

Table~\ref{table:deep_fashion} shows the results on \textit{mini}DeepFashion.
We observe that it is exceptionally challenging to reexamine distance ranking between instances online when the context of target class similarity switches from that of (meta-)training class.
CRML and the baselines result in moderate performance growth from DML, opening the door for future work.
Figure~\ref{fig:qual_fashion} shows the qualitative retrieval results of DML and CRML on the \textit{mini}DeepFashion. 
The leftmost images are queries and the right eight images are top-eight nearest neighbors. 
As shown in Fig.~\ref{fig:qual_fashion}, DML is misled by similar colors or shapes without adapting to target attributes, texture and category, only retrieving images of common patterns. 
In contrast, CRML adapts to attribute-specific data, thus retrieving correct images. 
For example, when the query is blue chinos, while DML only retrieves the blue pants regardless of the category, CRML retrieves the chinos successfully irrespective of the color (Fig.~\ref{fig:qual_fashion} (a)). 
Note that \mini DeepFashion benchmark has an original characteristics that requires online adaptation to a certain attribute of given a support set, thus this benchmark makes more sense for evaluating few-shot metric learning in comparison to prevalent few-shot classification benchmarks.

\section{Conclusion}
We have presented a few-shot metric learning framework and proposed an effective method, CRML, as well as other baseline methods.
All the few-shot metric learning methods consistently outperform the conventional metric learning approach, demonstrating that they effectively adapt the learned embedding using a few annotations from target classes. 
Moreover, few-shot metric learning is more effective than classification approaches on relational tasks such as learning with continuous labels and multi-attribute image retrieval tasks.
For this direction, we have introduced a challenging multi-attribute image retrieval dataset, \textit{mini}DeepFashion.
We believe few-shot metric learning is a new promising direction for metric learning, which effectively bridges the generalization gap of conventional metric learning. 

\subsubsection{Acknowledgements} 

This work was supported by Samsung Electronics Co., Ltd. (IO201208-07822-01), the IITP grants (2022-0-00959: Few-Shot Learning of Causal Inference in Vision and Language (30\%), 2019-0-01906: AI Graduate School Program at POSTECH (20\%)) and NRF grant (NRF-2021R1A2C3012728 (50\%)) funded by the Korea government (MSIT). 

%
%
%
%
\bibliographystyle{splncs04}
\bibliography{egbib}
\end{document}